%% file: final.tex
\documentclass[10pt,twocolumn,letterpaper]{article}

\usepackage{cvpr}
\usepackage{times}
\usepackage{epsfig}
\usepackage{graphicx}
\usepackage{amsmath}
\usepackage{amssymb}
\usepackage[misc]{ifsym}
\usepackage[bottom]{footmisc}
\usepackage[linesnumbered,ruled]{algorithm2e}

\usepackage{dblfloatfix}
\usepackage{float}

\usepackage[x11names,dvipsnames,table]{xcolor} 
\usepackage{diagbox}
\usepackage{multirow}
\usepackage{array}
\newcolumntype{L}[1]{>{\raggedright\let\newline\\\arraybackslash\hspace{0pt}}m{#1}}
\newcolumntype{C}[1]{>{\centering\let\newline\\\arraybackslash\hspace{0pt}}m{#1}}
\newcolumntype{R}[1]{>{\raggedleft\let\newline\\\arraybackslash\hspace{0pt}}m{#1}}

\usepackage[pagebackref=true,breaklinks=true,letterpaper=true,colorlinks,bookmarks=false]{hyperref}

\cvprfinalcopy 

\def\cvprPaperID{1070} 

\ifcvprfinal\fi
\begin{document}

\title{Elastic Boundary Projection for 3D Medical Image Segmentation}

\author{Tianwei Ni$^1$, Lingxi Xie$^{2,3(\textrm{\Letter})}$, Huangjie Zheng$^4$, Elliot K. Fishman$^5$, Alan L. Yuille$^2$\\
$^1$Peking University\quad$^2$Johns Hopkins University\quad$^3$Noah's Ark Lab, Huawei Inc.\\
$^4$Shanghai Jiao Tong University\quad$^5$Johns Hopkins Medical Institute\\
{\tt\small\{twni2016, 198808xc, alan.l.yuille\}@gmail.com}\quad{\tt\small zhj865265@sjtu.edu.cn}\quad{\tt\small efishman@jhmi.edu}
}

\maketitle
\thispagestyle{empty}

\begin{abstract}
We focus on an important yet challenging problem: using a 2D deep network to deal with 3D segmentation for medical image analysis. Existing approaches either applied multi-view planar (2D) networks or directly used volumetric (3D) networks for this purpose, but both of them are not ideal: 2D networks cannot capture 3D contexts effectively, and 3D networks are both memory-consuming and less stable arguably due to the lack of pre-trained models.

In this paper, we bridge the gap between 2D and 3D using a novel approach named Elastic Boundary Projection (EBP). The key observation is that, although the object is a 3D volume, what we really need in segmentation is to find its boundary which is a 2D surface. Therefore, we place a number of pivot points in the 3D space, and for each pivot, we determine its distance to the object boundary along a dense set of directions. This creates an elastic shell around each pivot which is initialized as a perfect sphere. We train a 2D deep network to determine whether each ending point falls within the object, and gradually adjust the shell so that it gradually converges to the actual shape of the boundary and thus achieves the goal of segmentation. EBP allows boundary-based segmentation without cutting a 3D volume into slices or patches, which stands out from conventional 2D and 3D approaches. EBP achieves promising accuracy in abdominal organ segmentation. Our code has been open-sourced \href{https://github.com/twni2016/Elastic-Boundary-Projection}{https://github.com/twni2016/EBP}.
\end{abstract}

\section{Introduction}
\label{Introduction}

Medical image analysis (MedIA), in particular 3D organ segmentation, is an important prerequisite of computer-assisted diagnosis (CAD), which implies a broad range of applications. Recent years, with the blooming development of deep learning, convolutional neural networks have been widely applied to this area~\cite{ronneberger2015u,milletari2016v}, which largely boosts the performance of conventional segmentation approaches based on handcrafted features~\cite{lin2006computer,ling2008hierarchical}, and even surpasses human-level accuracy in many organs and soft tissues.

Existing deep neural networks for medical image segmentation can be categorized into two types, differing from each other in the dimensionality of the processed object. The first type cuts the 3D volume into 2D slices, and trains a 2D network to deal with each slice either individually~\cite{yu2018recurrent} or sequentially~\cite{chen2016combining}. The second one instead trains a 3D network to deal with volumetric data directly~\cite{milletari2016v,liu20183d}. Although the latter was believed to have potentially a stronger ability to consider 3D contextual information, it suffers from two weaknesses: (1) the lack of pre-trained models makes the training process unstable and the parameters tuned in one organ less transferrable to others, and (2) the large memory consumption makes it difficult to receive the entire volume as input, yet fusing patch-wise prediction into the final volume remains non-trivial yet tricky.

\newcommand{\colwidthA}{1.14cm}
\newcommand{\colwidthB}{0.88cm}
\begin{table}
\small
\centering
{\setlength{\tabcolsep}{0.08cm}
\begin{tabular}{|l||C{\colwidthA}|C{\colwidthA}|C{\colwidthA}|C{\colwidthB}|}
\hline
{}                    &
    2D-Net \cite{ronneberger2015u,yu2018recurrent} &
    3D-Net \cite{milletari2016v,zhu20183d} &
    AH-Net \cite{liu20183d} &
    {\bf EBP} (ours) \\
\hline
Pure 2D network?      & \checkmark &            &            & \checkmark \\
\hline
Pure 3D network?      &            & \checkmark &            &            \\
\hline
Working on 3D data?   &            & \checkmark & \checkmark & \checkmark \\
\hline
3D data not cropped?  &            &            &            & \checkmark \\
\hline
3D data not rescaled? &            &            &            & \checkmark \\
\hline
Can be pre-trained?   & \checkmark &            & \checkmark & \checkmark \\
\hline
Segmentation method   & {\bf R}    & {\bf R}    & {\bf R}    & {\bf B}    \\
\hline
\end{tabular}}
\caption{A comparison between EBP and previous approaches in network dimensionality, data dimensionality, the ways of pre-processing data, network weights, and segmentation methodology. Due to space limit, we do not cite all related work here -- see Section~\ref{RelatedWork} for details. {\bf R} and {\bf B} in the last row stand for \textit{region-based} and \textit{boundary-based} approaches, respectively.}
\vspace{-0.2cm}
\label{Tab:ApproachComparison}
\end{table}

In this paper, we present a novel approach to bridge the gap between 2D networks and 3D segmentation. Our idea comes from the observation that an organ is often single-connected and locally smooth, so, instead of performing voxel-wise prediction, segmentation can be done by finding its boundary which is actually a 2D surface. Our approach is named Elastic Boundary Projection ({\bf EBP}), which uses the spherical coordinate system to project the irregular boundary into a rectangle, on which 2D networks can be applied. EBP starts with a pivot point within or without the target organ and a elastic shell around it. This shell, parameterized by the radius along different directions, is initialized as a perfect sphere (all radii are the same). The goal is to adjust the shell so that it eventually converges to the boundary of the target organ, for which we train a 2D network to predict whether each ending point lies inside or outside the organ, and correspondingly increase or decrease the radius at that direction. This is an iterative process, which terminates when the change of the shell is sufficiently small. In practice, we place a number of pivots in the 3D space, and summarize all the converged shells for outlier removal and 3D reconstruction.

Table~\ref{Tab:ApproachComparison} shows a comparison between EBP and previous 2D and 3D approaches. EBP enjoys three-fold advantages. First, EBP allows using a 2D network to perform volumetric segmentation, which absorbs both training stability and contextual information. Second, with small memory usage, EBP processes a 3D object entirely without cutting it into slices or patches, and thus prevents the trouble in fusing predictions. Third, EBP can sample abundant training cases by placing a number of pivots, which is especially useful in the scenarios of limited data annotations. We evaluate EBP in segmenting several organs in abdominal CT scans, and demonstrate its promising performance.

The remainder of this paper is organized as follows. Section~\ref{RelatedWork} briefly reviews related work, and Section~\ref{Approach} describes the proposed EBP algorithm. After experiments are shown in Section~\ref{Experiments}, we draw our conclusions in Section~\ref{Conclusions}.

\section{Related Work}
\label{RelatedWork}

Computer aided diagnosis (CAD) is a research area which aims at helping human doctors in clinics. Currently, a lot of CAD approaches start from medical image analysis to obtain accurate descriptions of the scanned organs, soft tissues, {\em etc.}. One of the most popular topics in this area is object segmentation, {\em i.e.}, determining which voxels belong to the target in 3D data, such as abdominal CT scans studied in this paper. Recently, the success of deep convolutional neural networks for image classification~\cite{krizhevsky2012imagenet,simonyan2015very,he2016deep,huang2017densely} has been transferred to object segmentation in both natural images~\cite{shelhamer2017fully,chen2018deeplab} and medical images~\cite{ronneberger2015u,milletari2016v}.

One of the most significant differences between natural and medical images lies in data dimensionality: natural images are planar (2D) while medical data such as CT and MRI scans are volumetric (3D). To deal with it, researchers proposed two major pipelines. The first one cut each 3D volume into 2D slices, and trained 2D networks to process each of them individually~\cite{ronneberger2015u}. Such methods often suffer from missing 3D contextual information, for which various techniques were adopted, such as using 2.5D data (stacking a few 2D images as different input channels)~\cite{roth2015deeporgan,roth2016spatial}, training deep networks from different viewpoints and fusing multi-view information at the final stage~\cite{zhou2017fixed,xia2018bridging,yu2018recurrent}, and applying a recurrent network to process sequential data~\cite{chen2016combining,cai2017pancreas}. The second one instead trained a 3D network to deal with volumetric data~\cite{cicek20163d,milletari2016v}. These approaches, while being able to see more information, often require much larger memory consumption, and so most existing methods worked on small patches~\cite{havaei2017brain,zhu20183d}, which left a final stage to fuse the output of all patches. In addition, unlike 2D networks that can borrow pre-trained models from natural image datasets~\cite{deng2009imagenet}, 3D networks were often trained from scratch, which often led to unstable convergence properties~\cite{tajbakhsh2016convolutional}. One possible solution is to decompose each 3D convolution into a 2D-followed-by-1D convolution~\cite{liu20183d}. A discussion on 2D vs. 3D models for medical image segmentation is available in~\cite{lai2015deep}.

Prior to the deep learning era, planar image segmentation algorithms were often designed to detect the boundary of a 2D object~\cite{hough1962method,ballard1981generalizing,boykov2001interactive,rother2004grabcut,liu2009paint}. Although these approaches have been significantly outperformed by deep neural networks in the area of medical image analysis~\cite{lin2006computer,ling2008hierarchical}, we borrow the idea of finding the 2D boundary instead of the 3D volume and design our approach.

\section{Elastic Boundary Projection}
\label{Approach}

\begin{figure*}
\centering
\includegraphics[width=17cm]{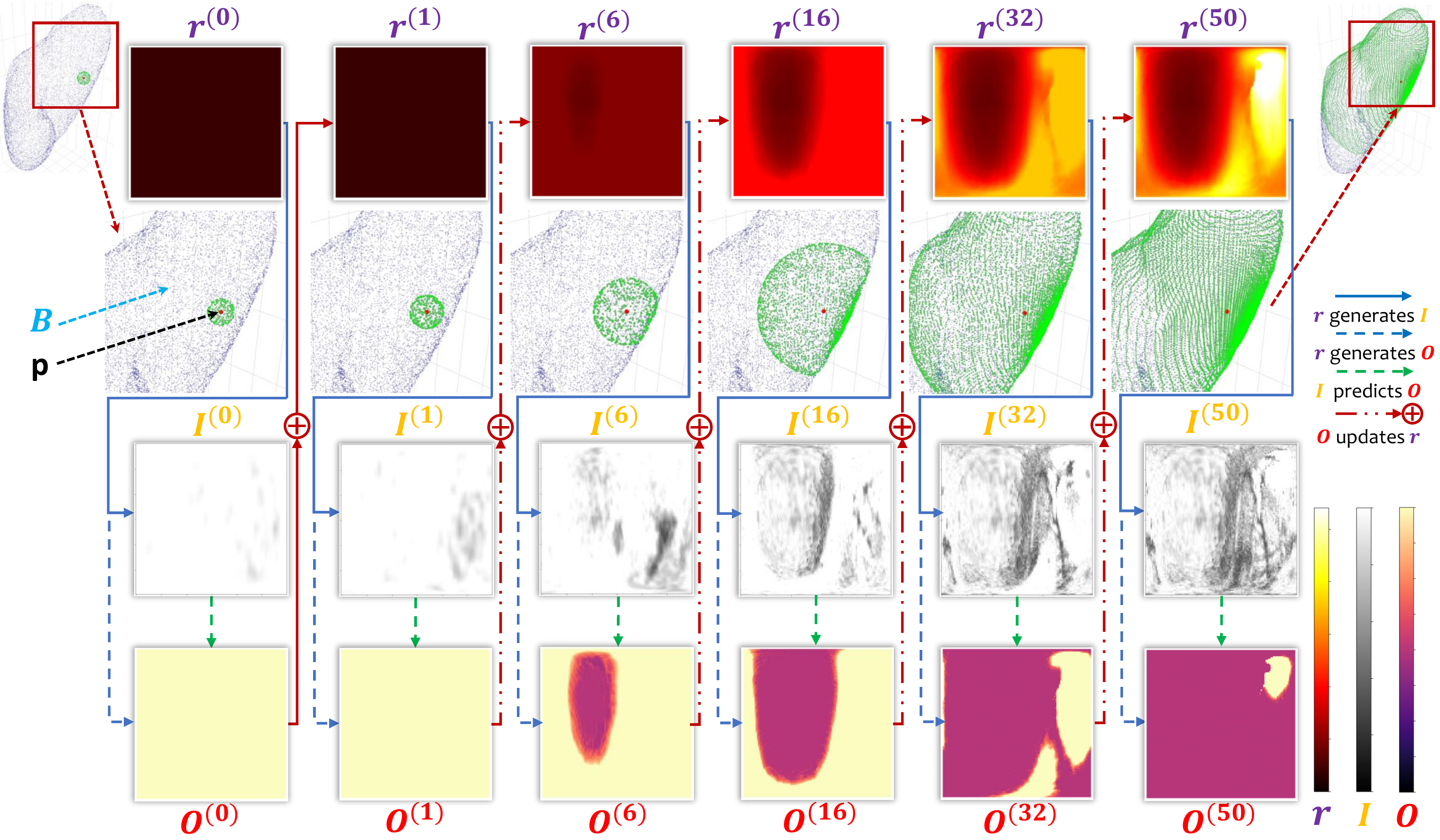}
\caption{
The overall flowchart of EBP (best viewed in color). We show the elastic shell after some specific numbers of iterations (green voxels in the second row) generated by a pivot $\mathbf{p}$ (the red center voxel in the second row) within a boundary $\mathcal{B}$ of the organ (blue voxels in 2nd row). The data generation process starts from a perfect sphere initialized by $r^{\left(0\right)}$, and then we obtain the $\left(\mathbf{I}^{\left(t\right)},\mathbf{O}^{\left(t\right)}\right)$ pairs (the third and fourth row) by $r^{\left(t\right)}$ in the training stage. In the testing stage, $\mathbf{O}^{\left(t\right)}$ is predicted by our model $\mathbb{M}$ given $\mathbf{I}^{\left(t\right)}$. After that, one iteration is completed by the adjustment of $r^{\left(t\right)}$ to $r^{\left(t+1\right)}$ by the addition of $\mathbf{O}^{\left(t\right)}$. Finally, the elastic shell converges to $\mathcal{B}$.
}
\vspace{-0.2cm}
\label{Fig:Flowchart}
\end{figure*}

\subsection{Problem, Existing Methods and Drawbacks}
\label{Approach:Motivation}

The problem we are interested in is to segment an organ from abdominal CT scans. Let an input image be $\mathbf{U}$, a 3D volume with $H_x\times H_y\times H_z$ voxels, and each voxel $U\!\left(x,y,z\right)$ indicates the intensity at the specified position measured by the Haunsfield unit (HU). The label $\mathbf{V}$ shares the same dimension with $\mathbf{U}$, and $V\!\left(x,y,z\right)$ indicates the class annotation of $U\!\left(x,y,z\right)$. Without loss of generality, we assume that ${V\!\left(x,y,z\right)}\in{\left\{0,1\right\}}$ where $1$ indicates the target organ and $0$ the background. Suppose our model predicts a volume $\mathbf{W}$, and ${\mathcal{V}}={\left\{\left(x,y,z\right)\mid V\!\left(x,y,z\right)=1\right\}}$ and ${\mathcal{W}}={\left\{\left(x,y,z\right)\mid W\!\left(x,y,z\right)=1\right\}}$ are the {\em foreground} voxels in ground-truth and prediction, respectively, we can compute segmentation accuracy using the Dice-S{\o}rensen coefficient (DSC): ${\mathrm{DSC}\!\left(\mathcal{V},\mathcal{W}\right)}={\frac{2\times\left|\mathcal{V}\cap\mathcal{W}\right|}{\left|\mathcal{V}\right|+\left|\mathcal{W}\right|}}$, which has a range of $\left[0,1\right]$ with $1$ implying a perfect prediction.

Let us denote the goal as  ${\mathbf{W}}={\mathbf{f}\!\left(\mathbf{U};\boldsymbol{\theta}\right)}$. Thus, there are two typical ways of designing $\mathbf{f}\!\left(\cdot;\boldsymbol{\theta}\right)$. The first one trains a 3D model to deal with volumetric data directly~\cite{cicek20163d,milletari2016v}, while the second one works by cutting the 3D volume into slices and using 2D networks for segmentation~\cite{roth2015deeporgan,yu2018recurrent}. Both 2D and 3D approaches have their advantages and disadvantages. We appreciate the ability of 3D networks to take volumetric cues into consideration (radiologists also exploit 3D information to make decisions), however, 3D networks are sometimes less stable, arguably because we need to train all weights from scratch, while the 2D networks can be initialized using pre-trained models from natural images ({\em e.g.}, RSTN~\cite{yu2018recurrent} borrowed FCN~\cite{long2015fully} as initialization). On the other hand, processing volumetric data ({\em e.g.}, 3D convolution) often requires heavier computation in both training and testing. We aim at designing an algorithm which takes both benefits of 2D and 3D approaches.

\subsection{EBP: the Overall Framework}
\label{Approach:Framework}

Our algorithm is named Elastic Boundary Projection (EBP). As the name shows, our core idea is to predict the boundary of an organ instead of every pixel in it.

Consider a binary volume $\mathbf{V}$ with $\mathcal{V}$ indicating the foreground voxel set. We define its {\em boundary} ${\mathcal{B}}={\partial\mathbf{V}}$ as a set of (continuous) coordinates that are located between the foreground and background voxels\footnote{The actual definition used in implementation is slightly different -- see Section~\ref{Approach:Training} for details.}. Since $\mathcal{B}$ is a 2D surface, we can parameterize it using a rectangle, and then apply a 2D deep network to solve it. We first define a set of {\em pivots} ${\mathcal{P}}={\left\{\mathbf{p}_1,\mathbf{p}_2,\ldots,\mathbf{p}_N\right\}}$ which are randomly sampled from the region-of-interest (ROI), {\em e.g.}, the 3D bounding-box of the object. Then, in the spherical coordinate system, we define a fixed set of {\em directions} ${\mathcal{D}}={\left\{\mathbf{d}_1,\mathbf{d}_2,\ldots,\mathbf{d}_M\right\}}$, in which each $\mathbf{d}_m$ is a unit vector $\left(\hat{x}_m,\hat{y}_m,\hat{z}_m\right)$, {\em i.e.}, ${\hat{x}_m^2+\hat{y}_m^2+\hat{z}_m^2}={1}$, for ${m}={1,2,\ldots,M}$. For each pair of $\mathbf{p}_n$ and $\mathbf{d}_m$, there is a {\em radius} $r_{n,m}$ indicating how far the boundary is along this direction, {\em i.e.}, ${\mathbf{e}_{n,m}}={\mathbf{p}_n+r_{n,m}\cdot\mathbf{d}_m}\in{\mathcal{B}}$\footnote{If $\mathbf{p}$ is located outside the boundary, there may exist some directions that the ray ${\mathbf{e}_{n,m}\!\left(r\right)}={\mathbf{p}_n+r\cdot\mathbf{d}_m}$ does not intersect with $\mathcal{B}$. In this case, we define ${r_{n,m}}={0}$, {\em i.e.}, along these directions, the boundary collapses to the pivot itself. In all other situations (including $\mathbf{p}$ is located within the boundary), there may be more than one $r_m$'s that satisfy this condition, in which cases we take the maximal $r_m$. When there is a sufficient number of pivots, the algorithm often reconstructs the entire boundary as expected. See Sections~\ref{Approach:Training} and~\ref{Approach:Testing} for implementation details.}. When $\mathcal{B}$ is not convex, it is possible that a single pivot cannot see the entire boundary, so we need multiple pivots to provide complementary information. Provided a sufficiently large number of pivots as well as a densely distributed direction set $\mathcal{D}$, we can approximate the boundary $\mathcal{B}$ and thus recover the volume $\mathcal{V}$ which achieves the goal of segmentation.

Therefore, volumetric segmentation reduces to the following problem: given a pivot $\mathbf{p}_n$ and a set of directions $\mathcal{D}$, determine all $r_{n,m}$ so that ${\mathbf{e}_{n,m}}={\mathbf{p}_n+r_{n,m}\cdot\mathbf{d}_m}\in{\mathcal{B}}$. This task is difficult to solve directly, which motivates us to consider the following counter problem: given $\mathbf{p}_n$, $\mathcal{D}$ and a group of $r_{n,m}$ values, determine whether these values correctly describe the boundary, {\em i.e.}, whether each $\mathbf{e}_{n,m}$ falls on the boundary. We train a model ${\mathbb{M}}:{\mathbf{O}}={\mathbf{f}\!\left(\mathbf{I};\boldsymbol{\theta}\right)}$ to achieve this goal. Here, the input is a generated image ${\mathbf{I}_{n}}\equiv{\left\{U\!\left(\mathbf{p}_n+r_{n,m}\cdot\mathbf{d}_m\right)\right\}_{m=1}^M}={ \left\{U(\mathbf{e}_{n,m})\right\}_{m=1}^M}$, where $U\!\left(\mathbf{e}_{n,m}\right)$ is the intensity value of $\mathbf{U}$ at position $\mathbf{e}_{n,m}$, interpolated by neighboring voxels if necessary. Note that $\mathbf{I}$ appears in a 2D rectangle. The output is a map $\mathbf{O}$ of the same size, with each value $o_m$ indicating whether $\mathbf{e}_{n,m}$ is located within, and how far it is from the boundary.

The overall flowchart of EBP is illustrated in Figure~\ref{Fig:Flowchart}. In the training stage, we sample $\mathcal{P}$ and generate $\left(\mathbf{I},\mathbf{O}\right)$ pairs to optimize $\boldsymbol{\theta}$. In the testing stage, we randomly sample $\mathcal{P}'$ and initialize all $r'_{n,m}$'s with a constant value, and use the trained model to iterate on each $\mathbf{p}'_n$ until convergence, {\em i.e.}, all entries in $\mathbf{O}'$ are close to $0$ (as we shall see later, convergence is required because one-time prediction can be inaccurate). Finally, we perform 3D reconstruction using all $\mathbf{e}'_{n,m}$'s to recover the volume $\mathbf{V}$. We will elaborate the details in the following subsections.

\subsection{Data Preparation: Distance to Boundary}
\label{Approach:DataPreparation}

In the preparation stage, based on a binary annotation $\mathbf{V}$, we aim at defining a relabeled matrix $\mathbf{C}$, with its each entry $C\!\left(x,y,z\right)$ storing the {\em signed distance} between each integer coordinate $\left(x,y,z\right)$ and $\partial\mathbf{V}$. The sign of $C\!\left(x,y,z\right)$ indicates whether $\left(x,y,z\right)$ is located within the boundary (positive: inside; negative: outside; $0$: on), and the absolute value indicates the distance between this point and the boundary (a point set). We follow the convention to define
\begin{equation}
\label{Eqn:BoundaryDistance}
{\left|C\!\left(x,y,z\right)\right|}={\min_{\left(x',y',z'\right)\in\partial\mathbf{V}}\mathrm{Dist}\!\left[\left(x,y,z\right),\left(x',y',z'\right)\right]},
\end{equation}
where we use the $\ell_2$-distance ${\mathrm{Dist}\!\left[\left(x,y,z\right),\left(x',y',z'\right)\right]}={\left(\left|x-x'\right|^2+\left|y-y'\right|^2+\left|z-z'\right|^2\right)^{1/2}}$ (the Euclidean distance) while a generalized $\ell_p$-distance can also be used. We apply the KD-tree algorithm for fast search. If other distances are used, {\em e.g.}, $\ell_1$-distance, we can apply other efficient algorithms, {\em e.g.}, floodfill, for constructing matrix $\mathbf{C}$. The overall computational cost is $O\!\left(N_0\log N_0^\circ\right)$, where ${N_0}={H_xH_yH_z}$ is the number of voxels and ${N_0^\circ}={\left|\partial\mathbf{V}\right|}$ is the size of the boundary set\footnote{Here are some technical details. The KD-tree is built on the set of {\em boundary voxels}, {\em i.e.}, the integer coordinates with at least one (out of six) neighborhood voxels having a different label (foreground vs. background) from itself. There are in average ${N_0^\circ}={50\rm{,}000}$ such voxels for each case, and performing $N_0$ individual searches on this KD-tree takes around $20$ minutes. To accelerate, we limit ${\left|C\!\left(x,y,z\right)\right|}\leqslant{\tau}$ which implies that all coordinates with a sufficiently large distance are truncated (this is actually more reasonable for training -- see the next subsection). We filter all pixels with an $\ell_{-\infty}$-distance not smaller than $\tau$\footnote{Mathematically, $\ell_{-\infty}$-distance (the minimal coordinate difference among three axes) is smaller than or equal to $\ell_2$-distance.}, which runs very fast\footnote{Using a modified version of the floodfill algorithm, we can find all these voxels in $O\!\left(N'\left\lceil\tau\right\rceil\right)$ time.} and typically reduces the number of searches to less than $1\%$ of $N_0$. Thus, data preparation takes less than $1$ minute for each case.}.

After $\mathbf{C}$ is computed, we multiply $C\!\left(x,y,z\right)$ by $-1$ for all background voxels, so that the sign of $C\!\left(x,y,z\right)$ distinguishes inner voxels from outer voxels. In the following parts, $\left(x,y,z\right)$ can be a floating point coordinate, in which case we use trilinear interpolation to obtain $C\!\left(x,y,z\right)$.

\subsection{Training: Data Generation and Optimization}
\label{Approach:Training}

To optimize the model $\mathbb{M}$, we need a set of training pairs $\left\{\left(\mathbf{I},\mathbf{O}\right)\right\}$. To maximally reduce the gap between training and testing data distributions, we simulate the iteration process in the training stage and sample data on the way.

We first define the direction set ${\mathcal{D}}={\left\{\mathbf{d}_1,\mathbf{d}_2,\ldots,\mathbf{d}_M\right\}}$. We use the spherical coordinate system, which means that each direction has an azimuth angle ${\alpha_{m_1}}\in{\left[0,2\pi\right)}$ and a polar angle ${\varphi_{m_2}}\in{\left[-\pi/2,\pi/2\right]}$. To organize these $M$ directions into a rectangle, we represent $\mathcal{D}$ as the Cartesian product of an azimuth angle set of $M^\mathrm{a}$ elements and a polar angle set of $M^\mathrm{p}$ elements where ${M^\mathrm{a}\times M^\mathrm{p}}={M}$. The $M^\mathrm{a}$ azimuth angles are uniformly distributed, {\em i.e.}, ${\alpha_{m_1}}={2m_1\pi/M^\mathrm{a}}$, but the $M^\mathrm{p}$ polar angles have a denser distribution near the equator, {\em i.e.}, ${\varphi_{m_2}}={\cos^{-1}\!\left(2m_2/\left(M^\mathrm{p}+1\right)-1\right)}-\pi/2$, so that the $M$ unit vectors are approximately uniformly distributed over the sphere. Thus, for each $m$, we can find the corresponing $m_1$ and $m_2$, and the unit direction vector $\left(\hat{x}_m,\hat{y}_m,\hat{z}_m\right)$ satisfies ${\hat{x}_m}={\cos\alpha_{m_1}\cos\varphi_{m_2}}$, ${\hat{y}_m}={\sin\alpha_{m_1}\cos\varphi_{m_2}}$ and ${\hat{z}_m}={\sin\varphi_{m_2}}$, respectively. ${\mathbf{d}_m}={\left(\hat{x}_m,\hat{y}_m,\hat{z}_m\right)}$. In practice, we fix ${M^\mathrm{a}}={M^\mathrm{p}}={120}$ which is a tradeoff between sampling density (closely related to accuracy) and computational costs.

We then sample a set of pivots ${\mathcal{P}}={\left\{\mathbf{p}_1,\mathbf{p}_2,\ldots,\mathbf{p}_N\right\}}$. At each $\mathbf{p}_n$, we construct a unit sphere with a radius of $R_0$, {\em i.e.}, ${r_{n,m}^{\left(0\right)}}={R_0}$ for all $m$, where the superscript $0$ indicates the number of undergone iterations. After the $t$-th iteration, the coordinate of each ending point is computed by:
\begin{equation}
\label{Eqn:EndingPoint}
{\mathbf{e}_{n,m}^{\left(t\right)}}={\mathbf{p}_n+r_{n,m}^{\left(t\right)}\cdot\mathbf{d}_m}.
\end{equation}
Using the coordinates of all $m$, we look up the ground-truth to obtain an input-output data pair:
\begin{equation}
\label{Eqn:InputOutput}
{\mathbf{I}_{n,m}^{\left(t\right)}}={U\!\left(\mathbf{e}_{n,m}^{\left(t\right)}\right)},\quad{\mathbf{O}_{n,m}^{\left(t\right)}}={C\!\left(\mathbf{e}_{n,m}^{\left(t\right)}\right)},
\end{equation}
and then adjust $r_{n,m}^{\left(t\right)}$ accordingly\footnote{Eqn~\ref{Eqn:DistanceIteration} is not strictly correct, because $\mathbf{d}_{m}$ is not guaranteed to be the fastest direction along which $\mathbf{e}_{n,m}^{\left(t\right)}$ goes to the nearest boundary. However, since $C\!\left(\mathbf{e}_{n,m}^{\left(t\right)}\right)$ is the shortest distance to the boundary, Eqn~\eqref{Eqn:DistanceIteration} does not change the inner-outer property of $\mathbf{e}_{n,m}^{\left(t\right)}$.}:
\begin{equation}
\label{Eqn:DistanceIteration}
{r_{n,m}^{\left(t+1\right)}}={\max\left\{r_{n,m}^{\left(t\right)}+C\!\left(\mathbf{e}_{n,m}^{\left(t\right)}\right),0\right\}},
\end{equation}
until convergence is achieved and thus all ending points fall on the boundary or collapse to $\mathbf{p}_n$ itself\footnote{If $\mathbf{p}_n$ is located within the boundary, then all ending points will eventually converge onto the boundary. Otherwise, along all directions with $\mathbf{e}_{n,m}^{\left(0\right)}$ being outside the boundary, $r_{n,m}^{\left(t\right)}$ will be gradually reduced to $0$ and thus the ending point collapses to $\mathbf{p}_n$ itself. These collapsed ending points will not be considered in 3D reconstruction (see Section~\ref{Approach:Reconstruction}).}.

When the 3D target is non-convex, there is a possibility that a ray ${\mathbf{e}_{n,m}\!\left(r\right)}={\mathbf{p}_n+r\cdot\mathbf{d}_m}$ has more than one intersections with the boundary. In this case, the algorithm will converge to the one that is closest to the initial sphere. We do not treat this issue specially in both training and testing, because we assume a good boundary can be recovered if (i) most ending points are close to the boundary and (ii) pivots are sufficiently dense.

Here we make an assumption: by looking at the projected image at the boundary, it is not accurate to predict that the radius along any direction should be increased or decreased by a distance larger than $\tau$ (we use ${\tau}={2}$ in experiments). So, we constrain ${C\!\left(x,y,z\right)}\in{\left[-\tau,\tau\right]}$. This brings three-fold benefits. First, the data generation process becomes much faster (see the previous subsection); second, iteration allows to generate more training data; third and the most important, this makes prediction easier and more reasonable, as we can only expect accurate prediction within a small neighborhood of the boundary.

After the training set is constructed, we optimize ${\mathbb{M}}:{\mathbf{O}}={\mathbf{f}\!\left(\mathbf{I};\boldsymbol{\theta}\right)}$ with regular methods, {\em e.g.}, stochastic gradient descent is used in this paper. Please see section \ref{Experiments:Settings} for the details of $\mathbb{M}$. As a side comment, our approach can generate abundant training data by increasing $N$ and thus the sampling density of pivots, which is especially useful when the labeled training set is very small.

\subsection{Testing: Iteration and Inference}
\label{Approach:Testing}

The testing stage is mostly similar to the training stage, which starts with a set of randomly placed pivots and a unit sphere around each of them. We fix the parameters $\boldsymbol{\theta}$ and iterate until convergence or the maximal number of rounds $T$ is reached (unlike training in which ground-truth is provided, iteration may not converge in testing). After that, all ending points of all pivots, except for those collapsed to the corresponding pivot, are collected and fed into the next stage, {\em i.e.}, 3D reconstruction. The following techniques are applied to improve testing accuracy.

{\bf First}, the input image $\mathbf{I}_n^{\left(t\right)}$ at each training/testing round only contains intensity values at the current shell defined by $\left\{\mathbf{e}_{n,m}^{\left(t\right)}\right\}$. However, such information is often insufficient to accurately predict $\mathbf{O}_n^{\left(t\right)}$, so we complement it by adding more {\em channels} to $\mathbf{I}_n^{\left(t\right)}$. The $l$-th channel is defined by $M$ radius values $\left\{s_{n,l,m}^{\left(t\right)}\right\}$. There are two types of channels, with $L^\mathrm{A}$ of them being used to sample the boundary and $L^\mathrm{B}$ of them to sample the inner volume:
\begin{equation}
\label{Eqn:ChannelSampling}
\begin{aligned}
{s_{n,l^\mathrm{A},m}^{\left(t\right)}} & ={r_{n,m}^{\left(t\right)}+l^\mathrm{A}-\left(L^\mathrm{A}+1\right)/2}, \\
{s_{n,l^\mathrm{B},m}^{\left(t\right)\prime}} & ={\frac{l^\mathrm{B}}{L^\mathrm{B}+1}\left[r_{n,m}^{\left(t\right)}-\left(L^\mathrm{A}+1\right)/2\right]}.
\end{aligned}
\end{equation}
When ${L^\mathrm{A}}={1}$ and ${L^\mathrm{B}}={0}$, it degenerates to using one single slice at the boundary. With relatively large $L^\mathrm{A}$ and $L^\mathrm{B}$ ({\em e.g.}, ${L^\mathrm{A}}={L^\mathrm{B}}={5}$ in our experiments), we benefit from seeing more contexts which is similar to volumetric segmentation but the network is still 2D. The number of channels in $\mathbf{O}$ remains to be $1$ regardless of $L^\mathrm{A}$ and $L^\mathrm{B}$.

{\bf Second}, we make use of the spatial consistency of distance prediction to improve accuracy. When the radius values at the current iteration $\left\{r_{n,m}^{\left(t\right)}\right\}$ are provided, we can randomly sample $M$ numbers ${\varepsilon_m}\sim{\mathcal{N}\!\left(0,\sigma^2\right)}$ where $\sigma$ is small, add them to $\left\{r_{n,m}^{\left(t\right)}\right\}$, and feed the noisy input to $\mathbb{M}$. By spatial consistency we mean the following approximation is always satisfied for each direction $m$:
\begin{equation}
\label{Eqn:SpatialConsistency}
{C\!\left(\mathbf{e}_{n,m}^{\left(t\right)}+\varepsilon_m\cdot\mathbf{d}_m\right)}={C\!\left(\mathbf{e}_{n,m}^{\left(t\right)}\right)+\varepsilon_m\cdot\cos\beta\!\left(\mathbf{d}_{m},\mathbf{e}_{n,m}^{\left(t\right)}\right)},
\end{equation}
where $\beta\!\left(\mathbf{d}_{m},\mathbf{e}_{n,m}^{\left(t\right)}\right)$ is the angle between $\mathbf{d}_{m}$ and the normal direction at $\mathbf{e}_{n,m}^{\left(t\right)}$. Although this angle is often difficult to compute, we can take the left-hand side of Eqn~\eqref{Eqn:SpatialConsistency} as a linear function of $\varepsilon_m$ and estimate its value at $0$ using multiple samples of $\varepsilon_m$. This technique behaves like data augmentation and improves the stability of testing.

\begin{figure*}
\centering
\includegraphics[width=17cm]{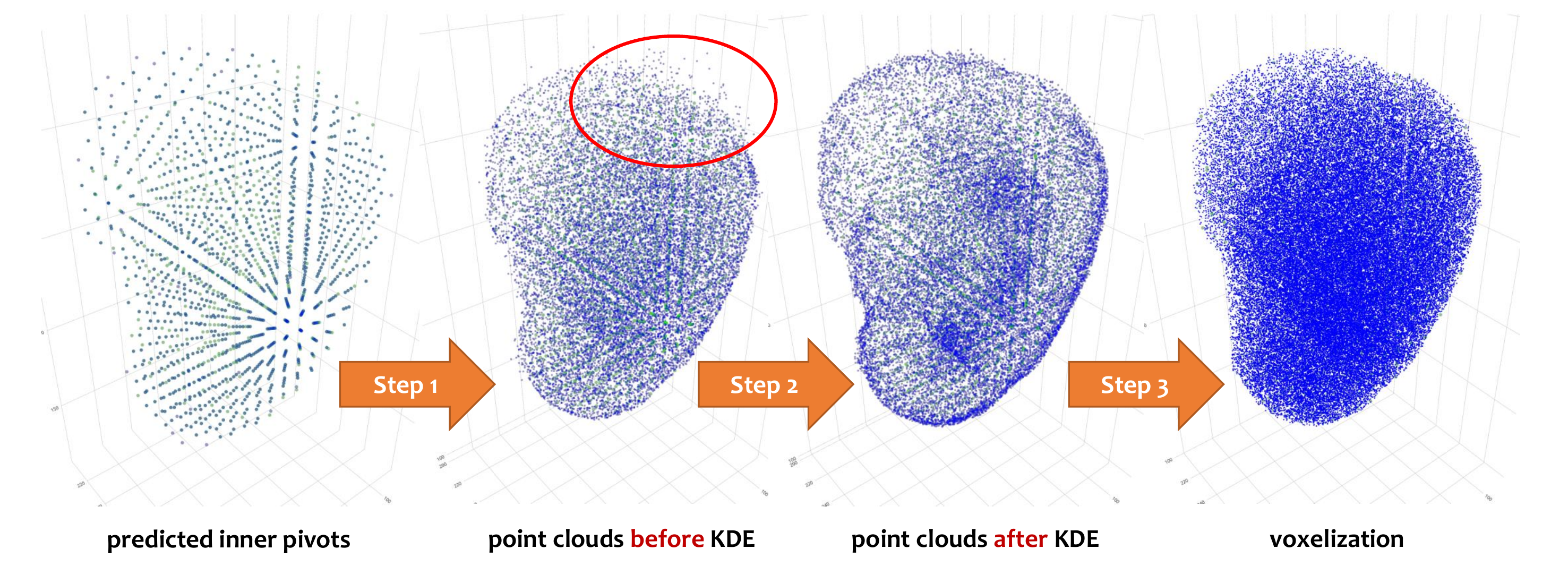}
\caption{
An example of 3D reconstruction (best viewed in color). We start with all pivots (green and blue points indicate ground-truth and predicted inner pivots, respectively) predicted to be located inside the target. In Step 1, all converged ending points generated by these pivots form the point clouds. In Step 2, a kernel density estimator (KDE) is applied to remove outliers (marked in a red oval in the second figure). In Step 3, we adopt a graphics algorithm for 3D reconstruction and finally we voxelize the point cloud.
}
\vspace{-0.2cm}
\label{Fig:Reconstruction}
\end{figure*}

{\bf Third}, we shrink the gap between training and testing data distributions. Note that in the training stage, all $r_{n,m}^{\left(t\right)}$ values are generated using ground-truth, while in the testing stage, they are accumulated by network predictions. Therefore, inaccuracy may accumulate with iteration if the model is never trained on such ``real'' data. To alleviate this issue, in the training stage, we gradually replace the added term in Eqn~\eqref{Eqn:DistanceIteration} with prediction, following the idea of curriculum learning~\cite{bengio2009curriculum}. In Figure~\ref{Fig:Flowchart}, we show that this strategy indeed improves accuracy in validation.

{\bf Last} but not least, we note that most false positives in the testing stage are generated by outer pivots, especially those pivots located within another organ with similar physical properties. In this case, the shell may converge to an unexpected boundary which harms segmentation. To alleviate this issue, we introduce an extra stage to determine which pivots are located within the target organ. This is achieved by constructing a graph with all pivots being nodes and edges being connected between neighboring pivots. The weight of each edge is the the intersection-over-union (IOU) rate between the two volumes defined by the elastic shells. To this end, so we randomly sample several thousand points in the region-of-interest (ROI) and compute whether they fall within each of the $N$ shells, based on which we can estimate the IOU of any pivot pairs. Then, we find the minimum cut which partitions the entire graph to two parts, and the inner part is considered the set of inner pivots. Only the ending points produced by inner pivots are considered in 3D reconstruction.

\subsection{3D Reconstruction}
\label{Approach:Reconstruction}

The final step is to reconstruct the surface of the 3D volume based on all ending points. Note that there always exist many false positives ({\em i.e.}, predicted ending points that do not fall on the actual boundary), so we adopt kernel density estimation (KDE) to remove them, based on the assumption that with a sufficient number of pivots, the density of ending points around the boundary is much larger than that in other regions. We use the Epanechnikov kernel with a bandwidth of $1$, and preserve all integer coordinates with a log-likelihood not smaller than $-14$.

Finally, we apply a basic graphics framework to accomplish this goal, which works as follows. We first use the Delaunay triangulation to build the mesh structure upon the survived ending points, and then remove improper tetrahedrons with a circumradius larger than $\alpha$. After we obtain the alpha shape, we use the subdivide algorithm to voxelize it into volumes with hole filling. Finally, we apply surface thinning to the volumes by $3$ slices. This guarantees a closed boundary, filling which obtains final segmentation. We illustrate an example of 3D reconstruction in Figure~\ref{Fig:Reconstruction}.

\subsection{Discussions and Relationship to Prior Work}
\label{Approach:Relationship}

The core contribution of EBP is to provide a 2D-based approach for 3D segmentation. To the best of our knowledge, this idea was not studied in the deep learning literature. Conventional segmentation approaches such as GraphCut~\cite{boykov2001interactive} and GrabCut~\cite{rother2004grabcut} converted 2D segmentation to find the minimal cut, a 1D contour that minimizes an objective function, which shared a similar idea with us. Instead of manually defining the loss function by using voxel-wise or patch-wise difference, EBP directly measures the loss with a guess and iteratively approaches the correct boundary. This is related to the {\em active contour} methods~\cite{kass1988snakes,chan2001active}.

In the perspective of dimension reduction, EBP adds a different solution to a large corpus of 2D segmentation approaches~\cite{ronneberger2015u,roth2015deeporgan,roth2016spatial,zhou2017fixed,yu2018recurrent} which cut 3D volumes into 2D slices without considering image semantics. Our solution enjoys the ability of extracting abundant training data, {\em i.e.}, we can sample from an infinite number of pivots (no need to have integer coordinates). This makes EBP stand out especially in the scenarios of fewer training data (see experiments). Also, compared to pure 3D approaches~\cite{cicek20163d,milletari2016v}, we provide a more efficient way of sampling voxels which reduces computational overheads as well as the number of parameters, and thus the risk of over-fitting.


\section{Experiments}
\label{Experiments}

\subsection{Datasets, Evaluation and Details}
\label{Experiments:Settings}

We evaluate EBP in a dataset with $48$ high-resolution CT scans. The width and height of each volume are both $512$, and the number of slices along the {\em axial} axis varies from $400$ to $1\rm{,}100$. These data were collected from some potential renal donors, and annotated by four expert radiologists in our team. Four abdominal organs were labeled, including {\em left kidney}, {\em right kidney} and {\em spleen}. Around $1$ hour is required for each scan. All annotations were later verified by an experienced board certified Abdominal Radiologist. We randomly choose half of these volumes for training, and use the remaining half for testing. The data split is identical for different organs. We compute DSC for each case individually, {\em i.e.}, ${\mathrm{DSC}\!\left(\mathcal{V},\mathcal{W}\right)}={\frac{2\times\left|\mathcal{V}\cap\mathcal{W}\right|}{\left|\mathcal{V}\right|+\left|\mathcal{W}\right|}}$ where $\mathcal{V}$ and $\mathcal{W}$ are ground-truth and prediction, respectively.

For the second dataset, we refer to the {\em spleen} subset in the Medical Segmentation Decathlon (MSD) dataset (website: {\small\sf http://medicaldecathlon.com/}). This is a public dataset with $41$ cases, in which we randomly choose $21$ for training and the remaining $20$ are used for testing. This dataset has quite a different property from ours, as the spatial resolution varies a lot. Although the width and height are still both $512$, the length can vary from $31$ to $168$. DSC is also used for accuracy computation.

Two recenty published baselines named RSTN~\cite{yu2018recurrent} and VNet~\cite{milletari2016v} are used for comparison. RSTN is a 2D-based network, which uses a coarse-to-fine pipeline with a saliency transformation module. We directly follow the implementation by the authors. VNet is a 3D-based network, which randomly crops into $128\times128\times64$ patches from the original patch for training, and uses a 3D sliding window followed by score average in the testing stage. Though RSTN does not require a 3D bounding-box (ROI) while EBP and VNet do, this is considered fair because a 3D bounding-box is relatively easy to obtain. In addition, we also evaluate RSTN with 3D bounding-box, and found very little improvement compared to the original RSTN.

The model ${\mathbb{M}}:{\mathbf{O}}={\mathbf{f}\!\left(\mathbf{I};\boldsymbol{\theta}\right)}$ of EBP is instantiated as a 2D neural network based on UNet~\cite{ronneberger2015u}. The input image $\mathbf{I}$ has a resolution of $M = {M^\mathrm{a}\times M^\mathrm{p}}= 120 \times 120$. We set ${L^\mathrm{A}}={L^\mathrm{B}}={5}$, and append $3$ channels of $\mathbf{d}$ for both parts (thus each part has $8$ channels, and group convolution is applied). Our network has $3$ down-sampling and $3$ up-sampling blocks, each of which has three consecutive $2$-group dilated (rate is $2$) convolutions. There are also short (intra-block) and long (inter-block) residual connections. The output $\mathbf{O}$ is a one-channel signed distance matrix.

\renewcommand{\colwidthA}{1.80cm}
\renewcommand{\colwidthB}{0.80cm}
\begin{table*}[!h]
\setlength{\tabcolsep}{0.06cm}
\small
\centering{
\begin{tabular}{|l||C{\colwidthA}|C{\colwidthB}|C{\colwidthB}||C{\colwidthA}|C{\colwidthB}|C{\colwidthB}||C{\colwidthA}|C{\colwidthB}|C{\colwidthB}||C{\colwidthA}|C{\colwidthB}|C{\colwidthB}|}
\hline
\multirow{2}{*}{Approach}   & \multicolumn{3}{c||}{{\em left kidney}}     & \multicolumn{3}{c||}{{\em right kidney}}
                            & \multicolumn{3}{c||}{{\em spleen}}          & \multicolumn{3}{c|}{{\bf MSD} {\em spleen}} \\
\cline{2-13}
{}                          & Average                 & Max     & Min     & Average                 & Max     & Min
                            & Average                 & Max     & Min     & Average                 & Max     & Min     \\
\hline\hline
RSTN~\cite{yu2018recurrent} &          $94.50\pm2.66$ & $97.69$ & $93.64$ &          $96.09\pm2.21$ & $98.18$ & $87.35$
                            &          $94.63\pm4.21$ & $97.38$ & $78.75$ &         $89.70\pm12.60$ & $97.25$ & $48.45$ \\
\hline
VNet~\cite{milletari2016v}  &          $91.95\pm4.63$ & $95.23$ & $71.40$ &          $92.97\pm3.67$ & $97.48$ & $80.51$
                            &          $92.68\pm3.25$ & $96.75$ & $83.18$ &          $92.94\pm3.58$ & $97.35$ & $81.96$ \\
\hline
{\bf EBP} (ours)            &          $93.45\pm1.62$ & $97.28$ & $90.88$ &          $95.26\pm1.59$ & $97.45$ & $90.19$
                            &          $94.50\pm2.64$ & $96.76$ & $89.67$ &          $92.01\pm4.50$ & $96.48$ & $77.07$ \\
\hline
\end{tabular}}
\caption{
Comparison of segmentation accuracy (DSC, $\%$) on our multi-organ dataset and the {\em spleen} class in the MSD benchmark. Within each group, average (with standard deviation), max and min accuracies are reported.
}
\vspace{-0.2cm}
\label{Tab:Results}
\end{table*}

\subsection{Quantitative Results}
\label{Experiments:QuantitativeResults}

Results are summarized in Table~\ref{Tab:Results}. In all these organs, EBP achieves comparable segmentation accuracy with RSTN, and usually significantly outperforms VNet. 

On our own data, EBP works slightly worse than RSTN, but on the {\em spleen} set, the worst case reported by RSTN has a much lower DSC ($78.75\%$) than that of EBP ($89.67\%$). After diagnosis, we find that RSTN fails to detect a part this organ in a few continuous 2D slices, but EBP, by detecting the boundary, successfully recovers this case. This suggests that in many cases, EBP and RSTN can provide supplementary information to organ segmentation. Moreover, on the MSD spleen dataset, a challenging public dataset, EBP outperforms RSTN by more than $2\%$. In addition, (i) the worst case in MSD {\em spleen} reported by EBP is $77.07\%$, much higher than $48.45\%$ reported by RSTN; (ii) all standard deviations reported by RSTN are significantly larger. Both the above suggest that EBP enjoys higher stability.

We can observe that VNet often suffers even lower stability in terms of both standard deviation and worst accuracy. In addition, the training process is not guaranteed to converge to a good model, {\em e.g.}, in {\em right kidney} of our own dataset, we trained two VNet models -- one of them, as shown in Table~\ref{Tab:Results}, is slightly worse than both RSTN and EBP; while the other, reports even worse results: $86.30\pm6.50\%$ average, $95.32\%$ max and $73.66\%$ min DSCs, respectively. Similar phenomena, which are mainly due to the difficulty of optimizing a 3D-based network, were also observed in~\cite{cicek20163d,liu20183d,xia2018bridging}.

We also observe the impact of spatial resolution in the MSD {\em spleen} dataset. This dataset has a relatively low spatial resolution ({\em i.e.}, $31$--$168$ voxels along the long axis), which raises extra difficulties to VNet (it requires $128\times128\times64$ patches to be sampled). To deal with this issue, we normalize all volumes so as to increase the number of slices along the long axis. The results of VNet shown in Table~\ref{Tab:Results} are computed in this normalized dataset (in DSC computation, all volumes are normalized back to the original resolution for fair comparison), while both RSTN and EBP directly work on the original non-normalized dataset. In addition, VNet reports an $71.07\pm36.27\%$ average DSC on the non-normalized dataset, with a few cases suffering severe false negatives. This implies that VNet heavily depends on data homogeneity, while EBP does not.

A complicated organ with irregular geometric shape and blurring boundary, \textit{pancreas}, is also investigated by these approaches. We find that voxel-wise/region-based methods such as RSTN perform well on it, with over $80\%$ accuracy on our own dataset. However, EBP predicts many false positives and thus only has about $60\%$ accuracy on the same setting. After careful diagnosis, we figure out that for pancreas segmentation, EBP tends to mistake some other organs around pancreas within 3D bounding-box, such as \textit{small bowel}, \textit{inferior vena cava} and \textit{duodenum}, for pancreas. There are several reasons accounting for the weakness in pancreas segmentation. \textbf{First}, those surrounding organs are intertwined with pancreas and some parts of their boundaries coincide within several voxels. \textbf{Second}, their boundaries look similar to that of pancreas from the perspective of intensity distribution, which adds difficulty to EBP. \textbf{Third}, 3D reconstruction of EBP is inaccurate for organs with irregular shape, for the difficulty in choosing the hyperparameter $\alpha$ to trim the convex hull of the predicted point cloud to the ground-truth irregular shape.

\subsection{How Does EBP Find the Boundary?}
\label{Experiments:Diagnosis}

Now, we discuss on how EBP finds the boundary. We study two aspects, {\em i.e.}, convergence and consistency, on one case of medium difficulty in the subset of {\em right kidney}.

\begin{figure}[!t]
\centering
\includegraphics[width=4cm]{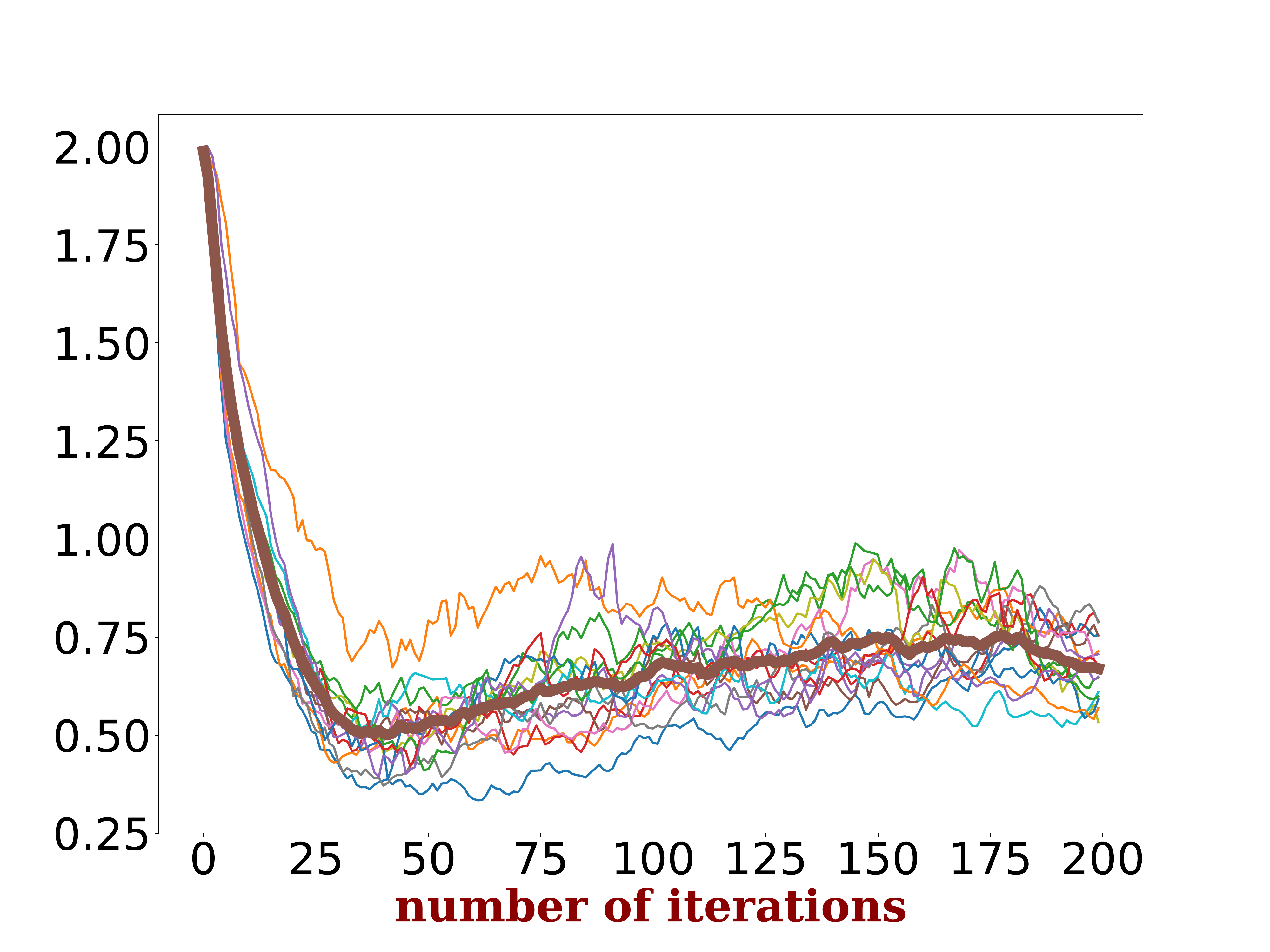}
\includegraphics[width=4cm]{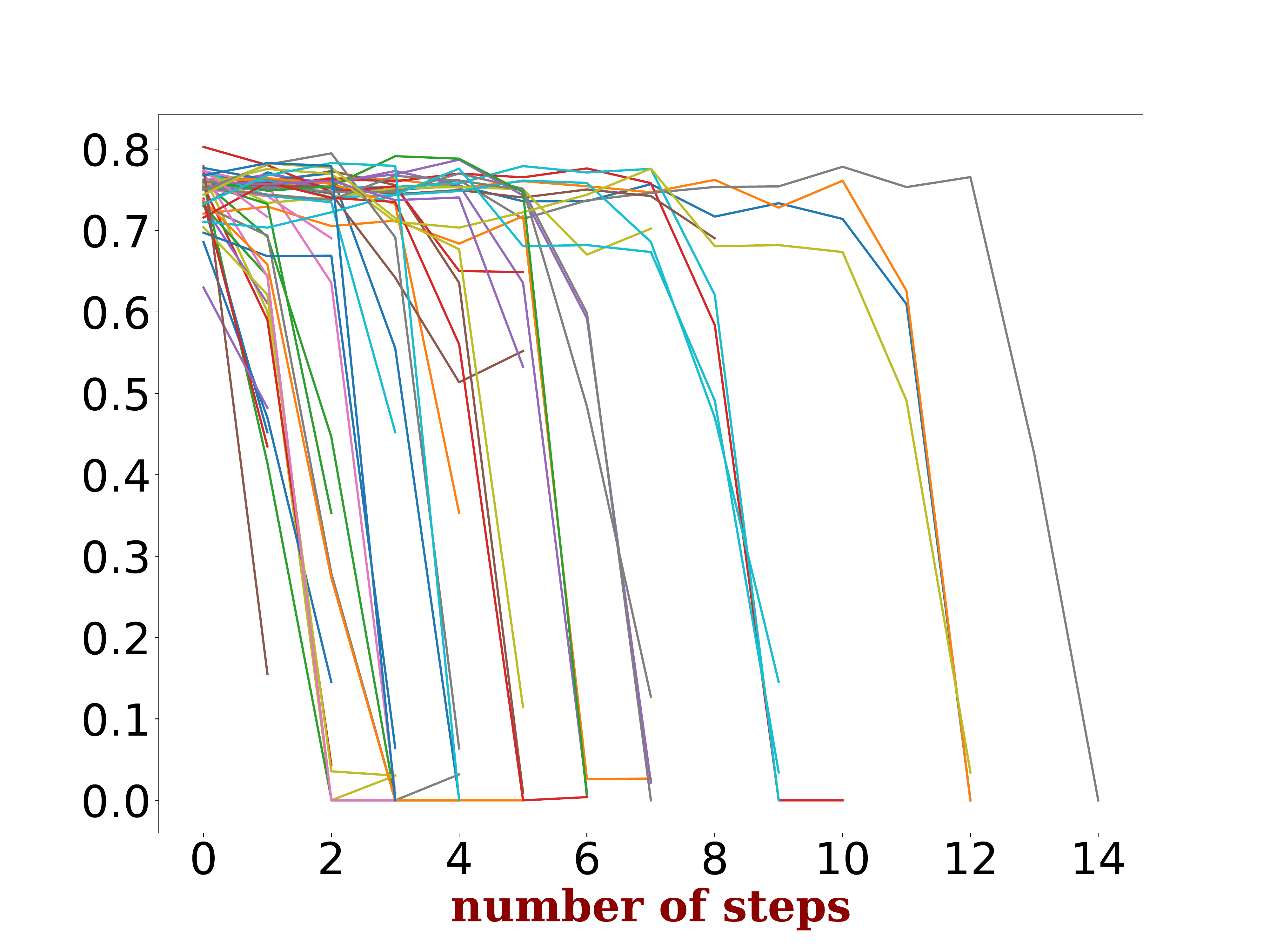}
\caption{
Left: the $\ell_1$-norm of $\mathbf{O}_{n,m}'$ during the first $200$ iterations. The thick curve is averaged over $15$ pivots, each of which appears as a thin curve. Right: The inter-pivot DSC recorded when a pivot keeps going along a fixed direction until it goes out of the target (we do not plot the curve beyond this point).
}
\vspace{-0.2cm}
\label{Fig:Properties}
\end{figure}

We start with investigating {\bf convergence}, by which we refer to whether each pivot $\mathbf{p}_n$, after a sufficient number of iterations, can converge to a boundary. We use the $\ell_1$-norm of $\mathbf{O}_{n,m}'$ to measure convergence, the output of which indicates the amount of revision along the radius. With its value reaching at a low level (positive but smaller than $0.5$), perfect convergence is achieved. Results are shown in Figure~\ref{Fig:Properties}. We can see that, starting from most inner pivots, the elastic shell can eventually converge to the boundary. In the figure, we show $200$ iterations, but in practice, for acceleration, we only perform $10$ iterations before sending all ``ending points'' to 3D reconstruction. This is to say, although convergence is not achieved and many ending points are not indeed located at the boundary, it is possible for 3D reconstruction algorithm to filter these outliers. This is because we have sampled a large number of pivots. Therefore, an ending point located near the boundary will be accompanied by a lot of others, while one located inside or even outside the target will be isolated. By applying kernel density estimation (KDE), we can filter out those isolated points so that 3D reconstruction is not impacted.

Next, we investigate {\bf consistency}, for which we take some pivot pairs and compute the DSC between the converged shells centered at them. This criterion was introduced in Section~\ref{Approach:Testing} to distinguish inner pivots from outer pivots. The assumption is that the shells generated by a pair of inner pivots should have a large similarity, while those generated by an inner pivot and an outer pivot should not. To maximally make fair comparison, we take all the {\em boundary pivots}, defined as the inner pivots with at least one neighbor being outside. Then, we sample all pivot pairs in which at least one of them is a boundary pivot, and make statistics. For those inner-inner pivot pairs, the average DSC ($73.46\%$) is much larger than that ($51.39\%$) of inner-outer pivot pairs. This experiment suggests that, two neighboring pivots are more likely to agree with each other if both of them are located within the target, otherwise the chance of getting a low DSC becomes large\footnote{There is a side note here. Theoretically, for an inner pivot and an outer pivot, if both elastic shells are perfectly generated, they should have a DSC of $0$. However, it is not often the case, because the elastic shell of the outer pivot is also initialized as a sphere, which may intersect with the boundary. In this case, all ending points that are initially located within the target will start growing until they reach the other border of the target. Consequently, it has a non-zero DSC with some of the inner pivots.}.

Last, we perform an interesting experiments to further reveal how inter-pivot DSC changes with the relative position of a pivot to the boundary. Starting from an inner pivot, we keep going along a fixed direction until being outside, and on the way, we record the DSC between the elastic shells generated by every neighboring pivot pairs. Some statistics are provided in Figure~\ref{Fig:Properties}. On all these curves, we observe a sudden drop at some place, which often indicates the moment that the pivot goes from inside to outside.


\section{Conclusions}
\label{Conclusions}

This paper presents EBP, a novel approach that trains 2D deep networks for 3D object segmentation. The core idea is to build up an elastic shell and adjust it until it converges to the actual boundary of the target. Since the shell is parameterized in the spherical coordinate system, we can apply 2D networks (low computational overhead, fewer parameters, pre-trained models, {\em etc.}) to deal with volumetric data (richer contextual information). Experiments are performed on several organs in abdominal CT scans, and EBP achieves comparable performance to both 2D and 3D competitors. In addition, EBP can sample sufficient training data from few annotated examples, which claims its advantage in medical image analysis.

We learn from this work that high-dimensional data often suffer redundancy ({\em e.g.}, not every voxel is useful in a 3D volume), and mining the discriminative part, though being challenging, often leads to a more efficient model. In the future, we will continue investigating this topic and try to cope with the weaknesses of EBP, so that it can be applied to a wider range of 3D vision problems, in particular when the object has a peculiar shape.

\vspace{0.4cm}
\noindent{\bf Acknowledgments}\quad This paper was supported by the Lustgarten Foundation for Pancreatic Cancer Research. We thank Prof. Zhouchen Lin for supporting our research. We thank Prof. Wei Shen, Dr. Yan Wang, Weichao Qiu, Zhuotun Zhu, Yuyin Zhou, Yingda Xia, Qihang Yu, Runtao Liu and Angtian Wang for instructive discussions.

{\small
\bibliographystyle{ieee}
\bibliography{egbib}
}

\input{supp}

\end{document}

%% file: supp.tex
\clearpage
\onecolumn

\section*{Appendix: Qualitative Results}
\label{Experiments}

\newcommand{\figurewidth}{16.0cm}
\begin{figure*}[!p]
\centering
\includegraphics[width=\figurewidth]{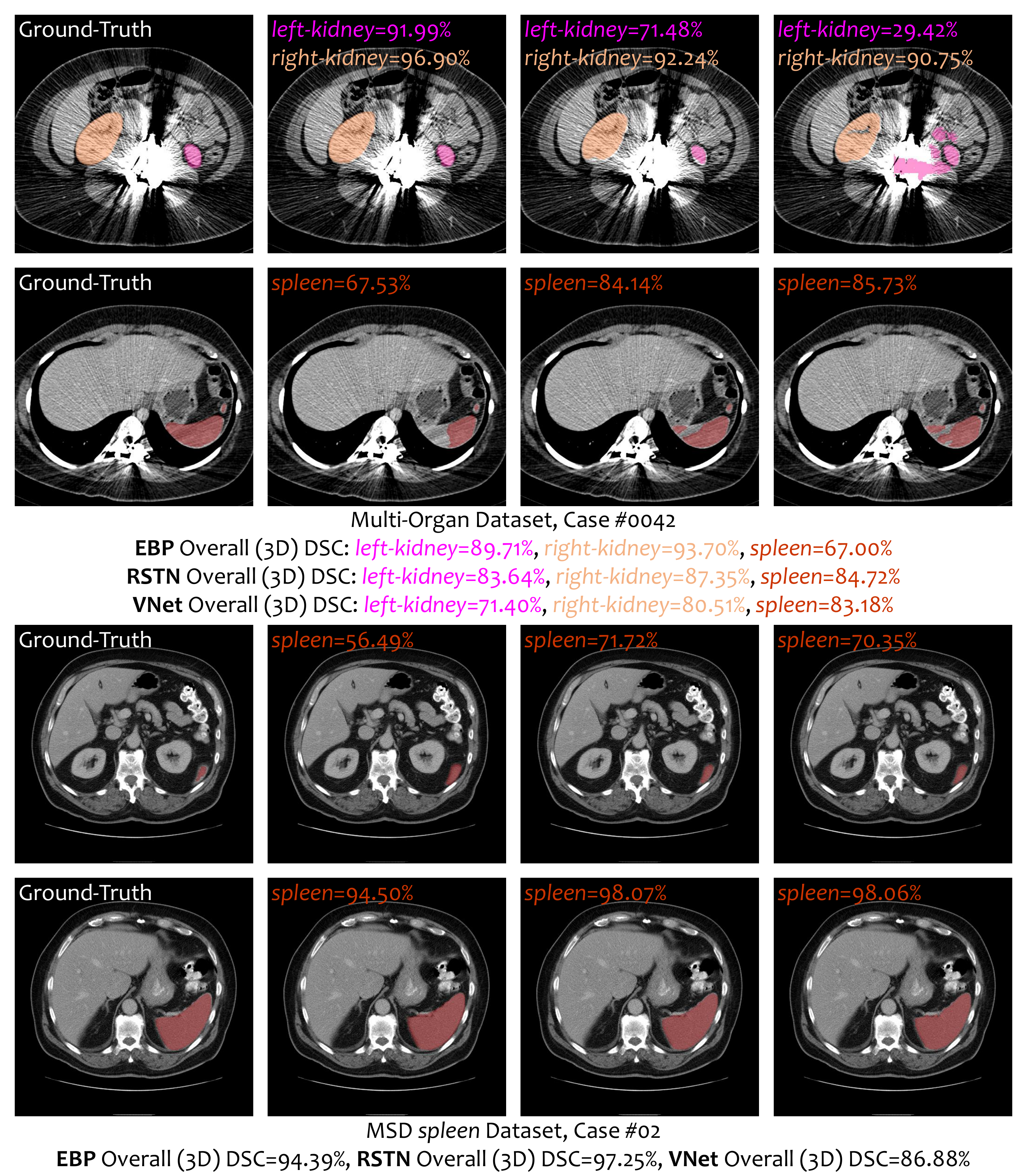}
\caption{2D visualization of segmentation results (best viewed in color). In each row, from left to right: ground-truth, EBP, RSTN, VNet. {\bf The top part} shows one special case in our multi-organ segmentation dataset. In this case, the image looks differently compared to most training images, due to some unusual situations during the CT scan. In this case, {\em kidney} segmentation results of both RSTN and VNet are heavily impacted whereas EBP works reasonably well. EBP produces unsatisfactory results on {\em spleen} segmentation, mainly because a part of pivots are not recognized as inner pivots. By simply tuning down the threshold by a little bit, EBP reports $86.44\%$ on {\em spleen} segmentation, which surpasses both RSTN and VNet. {\bf The bottom part} shows a case in the MSD {\em spleen} dataset, which we can observe how imperfect annotation affects DSC evaluations. In both rows, the ground-truth annotations do not cover the entire {\em spleen}. RSTN and VNet somehow miss a small margin close to the boundary, while EBP produces obviously better results but gets lower DSC scores. This tells us (i) ground-truth annotations in medical images are often imperfect; (ii) DSC values above the human level ({\em e.g.}, it can be defined as the average DSC between two individual human labelers, but such numbers are not available in most datasets) do not accurately reflect the absolute quality of segmentation, and in this scenario, a higher DSC does not guarantee better segmentation.}
\label{Fig:Visualization2D}
\end{figure*}

In the appendix, we visualize the segmentation results by RSTN~\cite{yu2018recurrent} (2D), VNet~\cite{milletari2016v} (3D) and EBP. The materials were not put in the main article due to the space limit. We choose one case from our dataset and one case from the MSD {\em spleen} dataset, respectively.

To compare the different behaviors between EBP and two baselines, we display some slice-wise segmentation results in Figure~\ref{Fig:Visualization2D}. We can see that EBP often produces results in a good shape, even when the image is impacted by some unusual conditions in CT scan. In comparison, RSTN and VNet produce segmentation by merging several parts (RSTN: slices, VNet: patches), therefore, in such extreme situations, some parts can be missing and thus segmentation accuracy can be low. On the other hand, the most common issue that harms the accuracy of EBP is the inaccuracy in distinguishing inner pivots from outer pivots. Under regular conditions, EBP is often more sensitive to the boundary of the targets, as it is especially trained to handle these cases -- an example comes from the visualization results in the MSD {\em spleen} dataset, which demonstrates that EBP sometimes produces better results than the ground-truth especially near the boundary areas.